\newcommand{\digits}{\text{Digits}}
\newcommand{\officecaltech}{\text{Office-Caltech}}

\newcommand{\mnist}{{MNIST}}
\newcommand{\usps}{{USPS}}
\newcommand{\svhn}{{SVHN}}
\newcommand{\syn}{{SYN}}
\newcommand{\amazon}{{Amazon}}
\newcommand{\caltech}{{Caltech}}
\newcommand{\dslr}{{DSLR}}
\newcommand{\webcam}{{Webcam}}

\newcommand{\resnett}{{ResNet-10}}

\newcommand{\fphl}{{FPHL}}
\newcommand{\fael}{{FAEL}}
\newcommand{\ours}{{FedHEAL}}
\newcommand{\fedavg}{{FedAvg}}
\newcommand{\qfedavg}{{q-FFL}}
\newcommand{\afl}{{AFL}}
\newcommand{\ditto}{{Ditto}}
\newcommand{\fedce}{{FedCE}}

\newcommand{\fedfv}{{FedFV}}
\newcommand{\fedprox}{{FedProx}}
\newcommand{\moon}{{MOON}}
\newcommand{\feddyn}{{FedDyn}}

\newcommand{\fedproc}{{FedProc}}
\newcommand{\fedproto}{{FedProto}}
\newcommand{\scaffold}{{Scaffold}}

\newcommand{\puc}{PUC}

\documentclass[10pt,twocolumn,letterpaper]{article}

\usepackage[pagenumbers]{cvpr} 

\usepackage[dvipsnames]{xcolor}

\usepackage{xcolor}
\usepackage{mathtools}
\definecolor{problemblue}{rgb}{0.6157, 0.7647, 0.9020}
\definecolor{problemyellow}{rgb}{1, 0.8510, 0.4}

\usepackage{bm}
\usepackage{multirow}
\usepackage[export]{adjustbox}
\newcommand{\myhyperlink}[3][black]{\hyperlink{#2}{\color{#1}{#3}}}
\usepackage{colortbl}
\definecolor{lightgray}{gray}{.9}
\definecolor{deepgray}{gray}{.8}

\usepackage{threeparttable}
\newcolumntype{I}{!{\vrule width 1pt}}
\makeatletter
\newcommand{\thickhline}{%
    \noalign {\ifnum 0=`}\fi \hrule height 1pt
    \futurelet \reserved@a \@xhline
}
\makeatother
\definecolor{lightgreen}{rgb}{0.06, 0.6, 0.06}
\definecolor{mygray}{gray}{.9}
\usepackage{float}
\usepackage[ruled, vlined]{algorithm2e}
\usepackage{pifont}
\usepackage{ragged2e}
\usepackage{enumitem}
\definecolor{cvprblue}{rgb}{0.21,0.49,0.74}
\usepackage[pagebackref,breaklinks,colorlinks,citecolor=cvprblue]{hyperref}

\def\paperID{4377} 
\def\confName{CVPR}
\def\confYear{2024}

\title{Fair Federated Learning under Domain Skew with \\Local Consistency and Domain Diversity} 

\author{
Yuhang Chen$^{1*}$ \quad Wenke Huang$^{1*}$ \quad Mang Ye$^{1,2\dagger}$\\
$^1$National Engineering Research Center for Multimedia Software,  \\
School of Computer Science, Wuhan University, Wuhan, China \\
$^2$Taikang Center for Life and Medical Sciences, Wuhan University, Wuhan, China \\
\texttt{\{yhchen0,wenkehuang,yemang\}@whu.edu.cn}\\
{\tt{\small{\url{https://github.com/yuhangchen0/FedHEAL}}}}
}

\begin{document}
\maketitle
\renewcommand{\thefootnote}{\fnsymbol{footnote}}
\footnotetext[1]{Equal contributions. $^{\dagger}$Corresponding author.}
\begin{abstract}
Federated learning (FL) has emerged as a new paradigm for privacy-preserving collaborative training. Under domain skew, the current FL approaches are biased and face two fairness problems. 1) Parameter Update Conflict: data disparity among clients leads to varying parameter importance and inconsistent update directions. These two disparities cause important parameters to potentially be overwhelmed by unimportant ones of dominant updates. It consequently results in significant performance decreases for lower-performing clients. 2) Model Aggregation Bias: existing FL approaches introduce unfair weight allocation and neglect domain diversity. It leads to biased model convergence objective and distinct performance among domains. We discover a pronounced directional update consistency in Federated Learning and propose a novel framework to tackle above issues. First, leveraging the discovered characteristic, we selectively discard unimportant parameter updates to prevent updates from clients with lower performance overwhelmed by unimportant parameters, resulting in fairer generalization performance. Second, we propose a fair aggregation objective to prevent global model bias towards some domains, ensuring that the global model continuously aligns with an unbiased model. The proposed method is generic and can be combined with other existing FL methods to enhance fairness. Comprehensive experiments on \digits{} and \officecaltech{} demonstrate the high fairness and performance of our method. 
\end{abstract}    
\section{Introduction}
\label{sec:intro}

\begin{figure}[t]
\centering
\includegraphics[trim=30 0 360 0, clip, width=\linewidth]{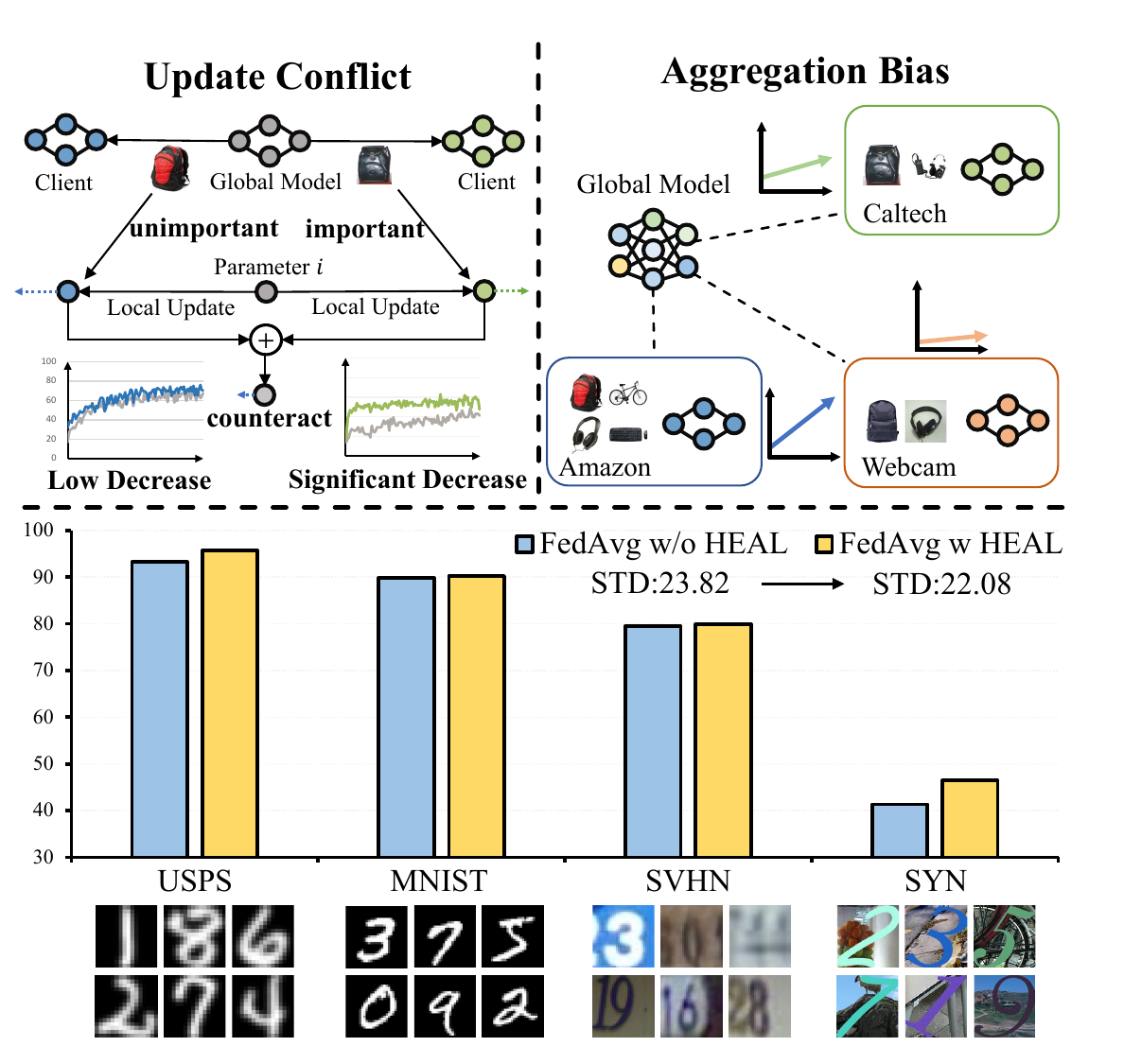}
\vspace{-20pt}
\captionsetup{font=small}
\caption{\small{
\textbf{Problem illustration} of Federated Learning under domain skew. Conventional FL methods (\fboxsep=0pt\fboxrule=0.6pt\fbox{\colorbox{problemblue}{\rule{0pt}{4pt}\rule{4pt}{0pt}}}) exhibit potential performance disparities due to Parameter Update Conflicts and Model Aggregation Bias. The former indicates that \textbf{varying parameter importance and inconsistent update directions} lead to an unfair decline in aggregated performance. The latter suggests biased convergence objective, resulting in \textbf{performance disparities}. Our method (\fboxsep=0pt\fboxrule=0.6pt\fbox{\colorbox{problemyellow}{\rule{0pt}{4pt}\rule{4pt}{0pt}}}) achieves more equitable performance across different domains while enhancing overall performance.
}}
\label{fig:problem}
\vspace{-15pt}
\end{figure}

Federated learning (FL) aims to collaboratively train a high-performance model while maintaining data privacy {\cite{FedAVG_AISTATS17_Communication,FLSurvey_SPM20_tianli}}. The foundational method, \fedavg{} {\cite{FedAVG_AISTATS17_Communication}}, allows numerous participants to send their models to the server instead of data. Then, the server aggregates these models into a global model and sends it back for further training. Notably, a significant challenge in FL is data heterogeneity {\cite{HFLSurvey_computingsurveys_Heterogeneous,FLSurvey_SPM20_tianli,FedProx_MLSys20_LossOptimization,SCAFFOLD_ICML20_stochastic,FedLSR_CIKM22}
}, which means that client data appears in a Non-IID (non-independently and identically distributed) manner {\cite{FLNoniid_arXiv18_noniid,FedIIC_MICCAI23,li2023no,AbrFun_SCIS2024,FedAS_CVPR24}. One particular heterogeneity type, domain skew {\cite{HFLSurvey_computingsurveys_Heterogeneous,FedBN_ICLR21_domainshift,FPL_CVPR23_Rethinking,FCCL_CVPR22_domainskew_modelhetero}}, refers to the client data being sampled from \textbf{various domains}, leading to different feature distributions for each client.

Under domain skew, the local data are sampled from multiple domains, resulting in a significant disparity in distributed data. This disparity introduces challenges of federated convergent inconsistency. Meanwhile, federated learning aims to achieve a lower overall loss {\cite{FedAVG_AISTATS17_Communication}}. These two factors lead to FL being biased toward domains with easier convergence, further resulting in the neglect of other domains. This bias leads to distinct performance among domains. 
However, if some clients feel undervalued, their motivation to participate in the federation will diminish, leading to a narrow scope of knowledge in the federation, hindering its growth and contradicting original intent of FL. 
This issue gives rise to a pivotal challenge in FL: \textbf{Performance Fairness} {\cite{FairFLSurvey_TNNLS23_fair,FLSurveyandBenchmarkforGenRobFair_arXiv23}}, which aims to ensure the uniform performance across different clients without neglecting clients with inferior performance. 
The fairness issue is highlighted in \cref{fig:problem}, where preliminary methods might overfit some domains, leading to poor performance in other domains. We argue that two primary reasons underlie this fairness issue: 
\hypertarget{Q1}{\textbf{\uppercase\expandafter{\romannumeral1}}} \textbf{Parameter Update Conflict}: \textit{The inconsistent parameter update directions and varying parameter importance lead to conflicts between important and unimportant parameters, degrading the performance of clients with poorer results.} Due to domain skew, there can be inconsistencies in parameter update directions among clients. Furthermore, some parameters of the neural network are more important to specific data {\cite{param_important_1989_LeCun,FedCurv_NeurIPSW19_CF, DynamicPFL_NeurIPS23}}, meaning that changes in these parameters have a larger impact on performance. Domain skew results in varying parameter importance. So important updates from poor-performing client may be potentially overwhelmed by unimportant aspects of others. But the latter can not signally boost performance. So it finally leads to performance disparity.
\hypertarget{Q2}{\textbf{\uppercase\expandafter{\romannumeral2}}} \textbf{Model Aggregation Bias}: \textit{The general weighting distribution method is biased and neglects domain diversity, resulting in unfair convergence objective and performance disparity.} In conventional FL methods {\cite{FedAVG_AISTATS17_Communication,FedProx_MLSys20_LossOptimization}}, the strategy of weighting proportional to sample quantity {\cite{FedAVG_AISTATS17_Communication}} hinders the global model from adequately learning from domains with few samples. Alternatively, equal weighting overly emphasizes clients with fewer samples. Both strategies introduce biases and amplifying performance diversity. This bias disregards the data diversity among different domains.

To address these issues, we present a novel solution, \textbf{Fed}erated Parameter-\textbf{H}armonized and Aggregation-\textbf{E}qu\textbf{A}lized \textbf{L}earning (\ours{}).
For problem \myhyperlink{Q1}{\textbf{\uppercase\expandafter{\romannumeral1}}}, we observe notable consistency in parameters updating during the local training. Specifically, due to the unique domain knowledge, some parameters are consistently pushed to the same direction (\ie, increment and decrement) during local training across consecutive rounds, as detailed in \cref{sec:fphl}.
Parameters with strong consistency occur because the global model fails to adapt to certain domains, leading to repeated adjustments in the same direction. Motivated by this, we argue that parameters with strong consistency are more crucial for specific domain. 
To mitigate parameter update conflict, we aim to prevent unimportant parameters from nullifying the crucial updates of domains with poor performance. By including only essential parameters, we mitigate important updates from being drowned out by less important ones, thus promoting Performance Fairness across multiple domains and clients.

For problem \myhyperlink{Q2}{\textbf{\uppercase\expandafter{\romannumeral2}}}, we argue that more diverse domains result in larger model changes during local training. To account for the data diversity, we propose an optimization objective that minimizes the variance of distances between the global model and all the local models. By reducing the variance, the global model maintains uniform distances to all the local models, preventing bias towards any clients.
However, the extensive parameters in neural networks and the large number of clients in FL make it computationally intensive. Thus, we propose a simplified method approximately aligned with this fairness objective. Details of our simplified approach are elaborated in \cref{sec:fael}.

In this paper, \ours{} consists of two components. \textbf{First}, by discarding unimportant parameters, we mitigate conflict among parameter updates. \textbf{Second}, we present a fair aggregation objective and a simplified implementation to prevent global model bias towards some domains. \ours{} ultimately achieves Performance Fairness under domain skew. Since our approach only focuses on aggregation, it can be easily integrated with most existing FL methods. Our main contributions are summarized as follows:

\begin{itemize}
    \item We identify the parameter update consistency in FL and introduce a partial parameter update method to update only parameters significant to the local domain, enhancing fair performance across domains.
    
    \item We propose a new fair federated aggregation objective and a practical approach to consider the domain diversity to improve Performance Fairness. 
    
    \item We conduct comprehensive experiments on the \digits{} {\cite{MNIST_IEEE98_digits,USPS_PAMI94_digits,SVHN_NeurIPS11_digits,SYN_arXiv18_digits}} and \officecaltech{} {\cite{OfficeCaltech_CVPR12_OC}} datasets, providing evidence of effectiveness of our method through ablation studies and integrations with existing methods.

\end{itemize}

\section{Related Work}

\subsection{Federated Learning with Data Heterogeneity}
Federated learning aims to collaboratively train models without centralizing data to protect privacy. The pioneering work, FedAvg {\cite{FedAVG_AISTATS17_Communication}}, trains a global model by aggregating participants' local model parameter.  However, FedAvg is primarily designed for homogeneous data, and its performance degrades under data heterogeneity. Numerous methods based on FedAvg have emerged to address data heterogeneity {\cite{RHFL_CVPR22_noisy,GradMA_CVPR23_gradient-memory,FGSSL_IJCAI23,AugHFL_ICCV23,FSMAFL_ACMMM22,FedClip_arXiv}}. FedProx {\cite{FedProx_MLSys20_LossOptimization}}, SCAFFOLD {\cite{SCAFFOLD_ICML20_stochastic}} and FedDyn {\cite{FedDyn_ICLR21_Regularization}} constrain local updates by adding penalty terms. FedProto {\cite{FedProto_AAAI22_PrototypeLoss}} and MOON {\cite{MOON_CVPR21_ContrastiveL}} enhance the alignment between client-side training at the feature level. However, these methods overlook the issue of domain skew, leading to diminished performance in multi-domain scenarios. Some methods have now been developed to address domain skew, such as FedBN {\cite{FedBN_ICLR21_domainshift}} and FCCL {\cite{FCCLPlus_TPAMI23}}. However, these methods focus on personalized models rather than shared models, the latter of which requires additional public datasets. FPL {\cite{FPL_CVPR23_Rethinking}} focuses on addressing domain skew but requires each client to upload high-level feature information, contradicting the privacy-preserving nature of FL. FedGA {\cite{FedGA_CVPR23_DG}} and FedDG {\cite{FedDG_CVPR21_FedDG}} focus on the problem of unseen domain generalization, but the former requires an additional validation set, and the latter involves transmitting data information among multiple clients, posing privacy leakage concerns. In this paper, \ours{} does not require any additional datasets or the transmission of additional signal information. We solely focus on the most fundamental transmitted information in FL: the model updates themselves, to extract the necessary information for enhancing Performance Fairness in multi-domain scenarios.

\begin{table}[t]\small
\centering
\vspace{5pt}
\resizebox{\linewidth}{!}{
    \renewcommand\arraystretch{1.0}
\begin{tabular}{clIcl}
\hline\thickhline
\rowcolor{lightgray}
Notation & Description & Notation & Description \\ \hline\hline

$m$ & Client Index &
$p_m$ & Client Weight \\

$i$ & Parameter Index &
$q_{m,i}$ & Parameter Weight \\

$M$ & Client Volume &
$l_{m,i}$ & Increment Proportion \\

$G$ & Parameter Volume &
$c_{m,i}$ & \puc{} of Parameter \\

$N_m$ & Sample Size &
$\Delta w_m^t$ & Model Update \\

$\mathcal{W}^t$ & Global Model &
$\Delta w_{m,i}^t$ & Parameter Update \\

$w^t_m$ & Client Model &
$d_m$ & Model Distance \\

$\tau$ & Importance Threshold &
$\beta$ & Update Momentum \\

\hline
\end{tabular}
}
\vspace{-8pt}
\caption{ \textbf{Notation table} of this paper. }
\label{tab:notation}
\vspace{-15pt}
\end{table}

\subsection{Fair Federated Learning}
The fairness in FL is currently of widespread interest {\cite{privacyFairFLSurvey_ComputingSurvey23,Fjord_NeurIPS21,FairVFL_NeurIPS22}}. The mainstream categorizations of federated fairness fall into three classes {\cite{FairFLSurvey_TNNLS23_fair,FedCE_CVPR23_Fair}}: Performance Fairness {\cite{AFL_ICML19_agnostic,qFedavg_ICLR19_fair,Ditto_ICML21_Ditto,FedCE_CVPR23_Fair}}, Individual/Group Fairness {\cite{FairFed_AAAI23_groupfair,FedFB_arXiv21_groupfair,FCFL_NeurIPS21_fair}}, and Collaborative Fairness {\cite{CFFL_xxxx20_collaborative,CollaborativeFairness_KDD21_Tutorial,FedCE_CVPR23_Fair,CGSV_NeurIPS21,CoreFed_NeurIPS22}}. Performance fairness ensures that all participants experience similar and equitable performance improvements. Individual/Group Fairness aims to minimize model bias towards specific attributes(\eg, gender). Collaborative Fairness ensures that participants are rewarded in proportion to contributions. This paper primarily addresses the issue of Performance Fairness. AFL {\cite{AFL_ICML19_agnostic}} utilizes a min-max optimization to boost the performance of the worst-performing clients. q-Fedavg {\cite{qFedavg_ICLR19_fair}} recalibrates the aggregate loss by assigning higher weights to devices with higher losses to enhance performance fairness. FedFV {\cite{FedFV_IJCAI21}} uses the cosine similarity to detect and eliminate gradient conflicts to achieve Performance Fairness. But they are tailored for label skew Non-IID data and do not consider domain skew. Ditto {\cite{Ditto_ICML21_Ditto}} enhances Performance Fairness by incorporating a penalty term but employs a personalized model instead of a shared model. FedCE {\cite{FedCE_CVPR23_Fair}} addresses both Performance Fairness and Collaborative Fairness but necessitates an additional validation set, a requirement that is challenging to meet given the scarcity of client data in FL. Our method is tailored for Performance Fairness under domain skew and consider both the domain diversity. It can be easily integrated with existing methods to enhance their fairness.
\section{Methodology}
\subsection{Preliminaries}\label{sec:pre}
\noindent \textbf{Federated Learning}. Following typical Federated Learning setup {\cite{FedAVG_AISTATS17_Communication,FedProx_MLSys20_LossOptimization,FedSeg_CVPR23}}, we consider there are $M$ clients (indexed by $m$). Each client holds private data $D_m = \{x_i, y_i\}_{i=1}^{N_m}$, where $N_m$ represents the data size of client $m$. The optimization objective of FL is to minimize global loss:
\begin{equation}\small
\setlength\abovedisplayskip{2pt} 
\setlength\belowdisplayskip{2pt}
    \begin{split}
    \min_w F(w) = \sum_{m=1}^M p_m f_m(w),
    \end{split}
    \label{eq:flobj}
\end{equation}
where \(f_m(w) = \frac{1}{N_m} \sum_{i=1}^{N_m} L(x_i, y_i; w)\), \(p_m\) is the weight of client. \(L(x_i, y_i; w)\) is the loss of data \((x_i, y_i)\) with model parameters \(w\). Each client updates its model locally, and the server then aggregates model updates from all clients.

\noindent \textbf{Domain Skew}. In heterogeneous federated learning, domain skew among private data occurs when the marginal distribution of labels \(P(y)\) is consistent across clients, but the conditional distribution of features given labels \(P(x|y)\) varies among different clients {\cite{FPL_CVPR23_Rethinking,FedBN_ICLR21_domainshift,FCCL_CVPR22_domainskew_modelhetero,advancesFL_FTML21,FedSM_CVPR22,FedCG_CVPRW21}}:
\begin{equation}\small
\setlength\abovedisplayskip{6pt} 
\setlength\belowdisplayskip{2pt}
P_m(x|y) \neq P_n(x|y) \quad \text{while} \quad P_m(y) = P_n(y).
\label{eq:domainskew}
\end{equation}

\subsection{Motivation}
\noindent \textbf{Observation of Parameter Update Consistency}. 
In heterogeneous federated learning, grasping the characteristic of model updates across clients is crucial. This study introduces a novel observation, termed as \textbf{Parameter Update Consistency} (\puc{}), observed during local training phases. Through a toy experiment involving 4 clients, each sampling from distinct domains and training with a \resnett{} network, we observe a significant Parameter Update Consistency during local training. As shown in \cref{fig:consistency}, our findings indicate that a substantial proportion of parameters maintain consistent update directions in consecutive rounds of training. Specifically, most parameters demonstrate significant update consistency. This consistency is noticeable in shorter consecutive rounds (10 rounds) but persists even in the longer term (100 rounds), underscoring its enduring nature. Furthermore, the consistency observed in the last 10 rounds reveals that \puc{} remains prominent as the global model converges, indicating that \textbf{the converged global model has not adapted to specific domains}. These observations suggest a potential global model bias towards other domains, emphasizing the necessity for strategies that mitigate such biases and ensure fairer model aggregation.

\begin{figure}[t]

\centering
\includegraphics[trim=30 260 510 10, clip, width=1\linewidth]{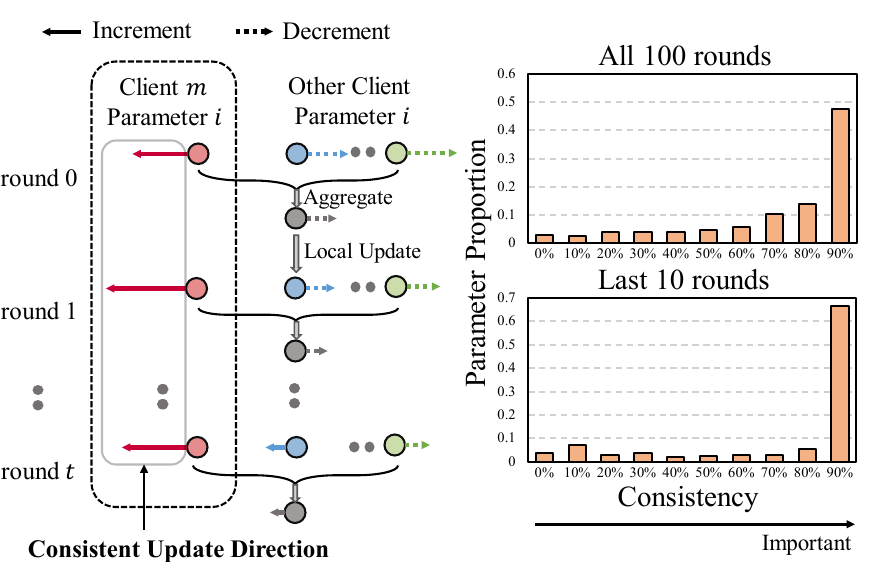}
\vspace{-10pt}
\captionsetup{font=small}
\caption{\small\textbf{Illustration of Parameter Update Consistency}. The consistency of parameter updates is displayed over 10 and 100 rounds for a randomly selected layer of the client model update. A significant proportion of parameters maintain a consistent update direction, \ie, almost half of the parameters show the same direction for over 90 of the 100 rounds, indicating a persistent tendency to steer the global model in a fixed direction.}
\label{fig:consistency}
\vspace{-15pt}
\end{figure}

\begin{figure*}[t]
	\begin{center}
		\includegraphics[trim=0 140 308 0, clip, width=0.9\linewidth]{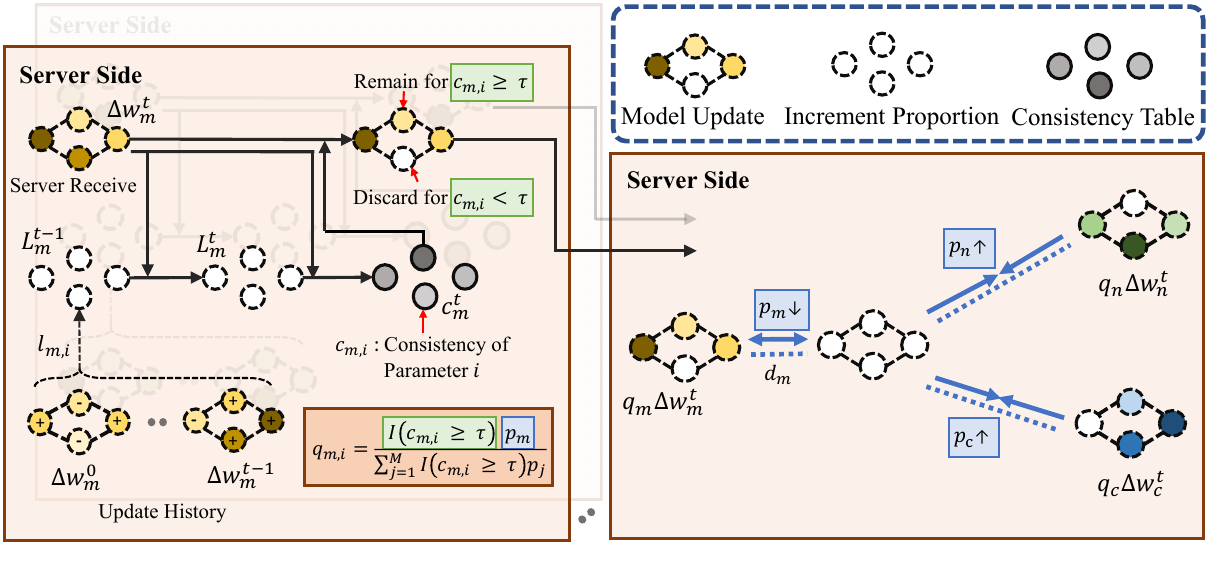}
            \put(-360,160){\scriptsize\cref{eq:new_weight}} 
            \put(-415,79){\scriptsize\cref{eq:incre_prop}} 
		\put(-397,98){\scriptsize\cref{eq:update_incr}} 
		\put(-332,98){\scriptsize\cref{eq:new_puc}} 
            \put(-180,55){\scriptsize Obtain $d_m$ by \cref{eq:new_dis}} 
            \put(-180,110){\scriptsize Update $p_m$ by \cref{eq:momentum}} 
            \put(-435,-5){(a) Federated Parameter-Harmonized Learning}
            \put(-207,-5){(b) Federated Aggregation-Equalized Learning}
	\end{center}
	\vspace{-15pt}
	\captionsetup{font=small}
	\caption{\small\textbf{Architecture illustration} of \ours{}. Clients send local model updates to the server. In \fphl{} (\cref{sec:fphl}), the server maintains a consistency table, computes the consistency of current updates with past directions and discards updates with low consistency. Then in \fael{} (\cref{sec:fael}), server minimized the variance of distance between global model and client model to mitigate model aggregation bias. }
	\label{fig:framework}
	\vspace{-5pt}
\end{figure*}
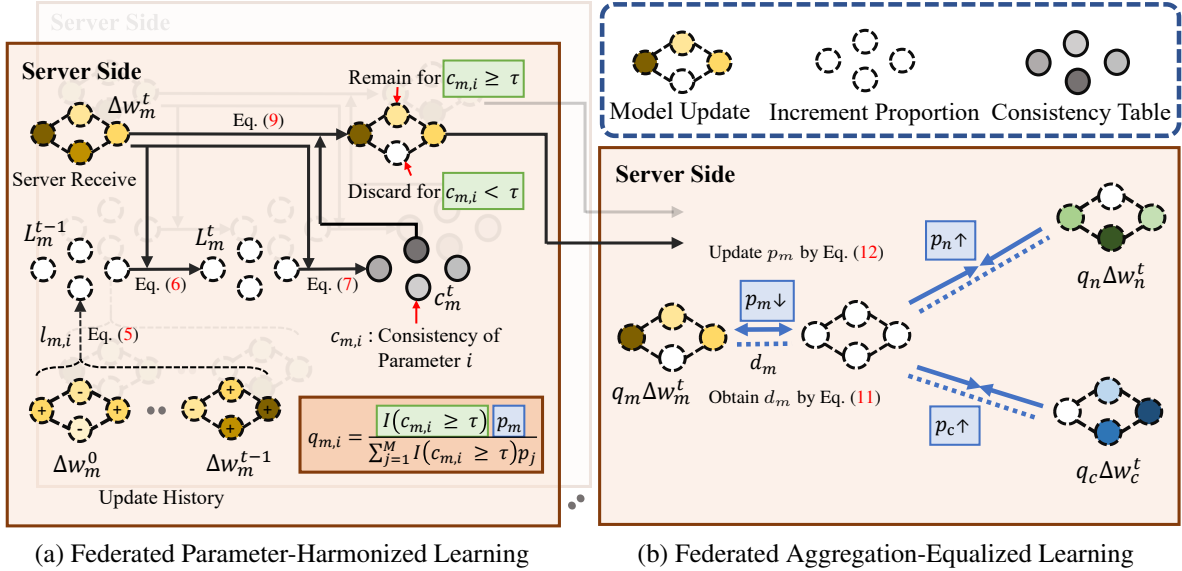
Inspired by our observations and insights, to tackle Problem \myhyperlink{Q1}{\textbf{\uppercase\expandafter{\romannumeral1}}}, we categorize parameters into two distinct classes: important and unimportant. Important parameters are characterized by stable update directions in consecutive rounds, indicating consistent learning behavior within specific domains and inadequacy of the global model in fitting those domains. These parameters are deemed crucial for offering greater contribution in performance improvement. Conversely, unimportant parameters exhibit distinct directional changes across rounds which provide little contribution in performance improvement. We argue that involving unimportant parameters in the updates exacerbates the parameter update conflict, leading to a decline in the performance of some domains and resulting in unfair performance. Therefore, to mitigate this, we aim to minimize the impact of unimportant updates, enhancing the performance of poorly performing domains. During parameter aggregation, we discard unimportant parameters by setting their weights to zero, which prevents them from participating in aggregation and mitigates the parameter update conflict. Then we normalize all weights of a local model parameter. Those with a weight of zero remain at zero but the weights of important parameters increase, amplifying their influence on the global model and improving its adaptation to underperforming domains. The proposed method is detailed in \cref{sec:fphl}.

To address Problem \myhyperlink{Q2}{\textbf{\uppercase\expandafter{\romannumeral2}}}, we employ domain diversity as the guiding metric for allocating weights. Our intuition is that more diverse data will undergo larger parameter changes for the local model to adapt to that domain. The parameter changes exhibit the fitting gap between the global model and local model, implying the potential for performance improvement. A larger model changes suggests a domain is overlooked and has greater space for performance enhancement, which can also be reflected in the magnitude of model parameter updates. The distance between model iterations can represent the extent of parameter updates. Thus, this issue can be addressed through an optimization objective that minimizes the variance of distances between the global model and each client model. By reducing the variance, the global model maintains a more uniform distance to each client. However, this optimization problem is computationally intensive, so we introduce a simplified approach. We modify the weight allocation strategy with a momentum update {\cite{Adam_ICLR14,ADAM_important_ICML13}}, assign greater weight to clients that induce larger parameter changes. This method approximately aligns with our objective and draws the new global model closer to neglected clients, thereby reducing the variance. We elaborate the proposed method in \cref{sec:fael}.

\subsection{Proposed Method}
\subsubsection{Federated Parameter-Harmonized Learning}\label{sec:fphl}
In a Federated Learning system with $M$ clients, the global model in round $t$ is denoted as $\mathcal{W}^t$. In round $t$, each client trains their model on private data $D_m$ to obtain local model $w_{m}^{t}$. The local model change is defined as $\Delta w_m^t = w_{m}^{t} - \mathcal{W}^t$. The global model of next round, $t+1$, is then updated by aggregating the model updates of round $t$:
\begin{equation}\small
\setlength\abovedisplayskip{2pt} 
\setlength\belowdisplayskip{2pt}
\mathcal{W}^{t+1} = \mathcal{W}^t + \sum_{m=1}^M p_m \Delta w_m^t,
\label{eq:rewri_agg}
\end{equation}
where $p_m$ is the aggregation weight of client $m$, typically proportion to local sample size(\ie{}, $p_m=\frac{N_m}{\sum_{j=1}^M N_j}$). In neural networks, $w_m^t$ is potentially a large vector of parameters, and $\Delta w_m^t$ is a vector of parameter changes with the same dimension. For simplicity, we disregard the internal structure of the model and represent $\Delta w_m^t$ as 
\begin{equation}\small
\setlength\abovedisplayskip{4pt} 
\setlength\belowdisplayskip{4pt}
\Delta w_m^t = [\Delta w_{m,1}^t, \Delta w_{m,2}^t, \ldots, \Delta w_{m,G}^t],
\label{eq:param}
\end{equation}
where $G$ denotes the total number of parameters, and $\Delta w_{m,i}^t$ represents the change in the $i^{th}$ parameter of the model $w_m^t$. 
We aim to compute the \puc{} for each parameter across all clients. This computation characterizes how significantly the parameters of the global model are pushed towards a consistent direction to fit specific domains.
We maintain a list of proportions for parameters that are increasing for each client and then employ dynamic programming {\cite{dynamic_programming}} to update this list in each round. The list of increment proportions for client $m$ is denoted as
\begin{equation}\small
\setlength\abovedisplayskip{2pt} 
\setlength\belowdisplayskip{2pt}
\begin{split}
L_m &= [l_{m,1},l_{m,2},\ldots,l_{m,G}], \\
l_{m,i} &= \frac{\sum_{j=0}^{t-1} I(\Delta w_{m,i}^j \geq 0)}{t},
\end{split}
\label{eq:incre_prop}
\end{equation}
where $I(\cdot)$ is the indicator function and $l_{m,i}$ is the proportion of increasing parameter before round $t$. Applying Dynamic Programming {\cite{dynamic_programming}}, the $l_{m,i}$ in the round $t$ can be calculated from the previous round’s result:
\begin{equation}\small
\setlength\abovedisplayskip{2pt} 
\setlength\belowdisplayskip{2pt}
l_{m,i} \leftarrow \frac{l_{m,i} * (t-1) + I(\Delta w_{m,i}^t \geq 0)}{t}.
\label{eq:update_incr}
\end{equation}
Then the \puc{} of $i^{th}$ parameter can be expressed as
\begin{equation}\small
\setlength\abovedisplayskip{2pt} 
\setlength\belowdisplayskip{2pt}
c_{m,i} = \puc{}(w_{m,i}) = 
\begin{cases} 
l_{m,i} & \text{if } \Delta w_{m,i}^t \geq 0, \\
1 - l_{m,i} & \text{otherwise}
\end{cases}.
\label{eq:new_puc}
\end{equation}
To alleviate parameter update conflicts and mitigate the global model from experiencing unfair performance, we select parameters with a strong \puc{} for retention and discard others. We introduce a hyperparameter $\tau$ for this purpose. We categorize the $i^{th}$ parameter as important if $c_{m,i} \geq \tau$ and as unimportant otherwise. 
Given the model aggregation process in \cref{eq:rewri_agg}, the aggregation of $i^{th}$ parameter is 
\begin{equation}\small
\setlength\abovedisplayskip{2pt} 
\setlength\belowdisplayskip{2pt}
\mathcal{W}_{i}^{t+1} = \mathcal{W}_{i}^{t} + \sum_{m=1}^M p_m \Delta w_{m,i}^t,
\label{eq:origin_param_agg}
\end{equation}
where $\mathcal{W}_{i}^{t}$ is the $i^{th}$ parameter of global model $\mathcal{W}^{t}$. By applying our method, \cref{eq:origin_param_agg} can be reformulated as
\begin{equation}\small
\setlength\abovedisplayskip{2pt} 
\setlength\belowdisplayskip{3pt}
\begin{aligned}
    \mathcal{W}_{i}^{t+1} &= \mathcal{W}_{i}^t + \sum_{m=1}^M q_{m,i}^t \Delta w_{m,i}^t, \\
    q_{m,i}^t &= \frac{I(c_{m,i} \geq \tau)p_m}{\sum_{j=1}^MI(c_{j,i} \geq \tau)p_j},
\end{aligned}
\label{eq:new_weight}
\end{equation}
where $p_m$ is the aggregation weight of client $m$ and all its parameters, $q_{m,i}^t$ zeros out the weights of insignificant parameters and then further normalizes them to ensure the aggregation weights sum up to 1. Consequently, only important parameters will participate and unimportant updates no longer impact important updates. By normalizing $q_{m,i}^t$, the proportion of important parameter updates is further amplified, enhancing their contribution during aggregation.

\subsubsection{Federated Aggregation-Equalized Learning}\label{sec:fael}

We first define the distance between the global model and the local model of client $m$ in round $t$ as $d_m = \left\| U - w_{m}^{t} \right\|_2^2$, where $U$ is the new global model and $\left\|\cdot\right\|_2^2$ is the square of the euclidean distance. To minimize the variance of distances between the global model and each client model, we introduce the following optimization objective:
\begin{equation}\small
\setlength\abovedisplayskip{1pt}
\setlength\belowdisplayskip{1pt}
\begin{aligned}
    U_t^* = &\arg \min_{U} \text{Var}(\{d_m\}_{m=1}^M) \\
    = &\arg \min_{U} \text{Var}(\{\left\| U - w_{m}^{t} \right\|_2^2\}_{m=1}^M) \\
    \text{s.t. } U = &\sum_{m=1}^{M} p_m w_{m}^{t}, \sum_{m=1}^{M} p_m = 1, \text{ and }  \forall m, p_m \geq 0,
\end{aligned}
\end{equation}
where \(U_t^*\) represents the unbiased global model. The time complexity of this optimization is $\mathcal{O}(qMG)$, where $q$ is the number of iterations needed for convergence. However, computational resources are often limited in FL. Thus, we propose a simplified approach that reduces the time complexity to $\mathcal{O}(MG)$, requiring only a single distance calculation for each client. Specifically, in round $t$, we measure the distance between the trained client model and global model $\mathcal{W}^t$ as $d_{m} = ||\Delta w_m^{t}||_2^2$. Notably, if we combine it with \fphl{}, $d_m$ can be rewritten as
\begin{equation}\small
\setlength\abovedisplayskip{1pt} \setlength\belowdisplayskip{1pt}
d_{m} = ||\sum_{i=1}^GI(c_{m,i} \geq \tau) \cdot \Delta w_{m,i}^t||_2^2,
\label{eq:new_dis}
\end{equation}
implying that we only compute the important parameter change distance which better reflects the alterations made for specific domain. We then apply a momentum update strategy {\cite{Adam_ICLR14,ADAM_important_ICML13}} to update the weight for each client:
\begin{equation}\small
\setlength\abovedisplayskip{0pt} \setlength\belowdisplayskip{0pt}
\begin{split}
\Delta p_m^t & = (1-\beta)\Delta p_m^{t-1} + \beta\frac{d_m}{\sum_{j=1}^Md_j}, \\
p_m^t & = p_m^{t-1} + \Delta p_m^t, \quad p_m^t = \frac{p_m^t}{\sum_{j=1}^M p_j^t}.
\end{split}
\label{eq:momentum}
\end{equation}
where $\beta$ is a hyper-parameter, with larger values indicating a more pronounced influence of the distance on $p_m$. When $\beta=0$, the method degenerates to the \fedavg{}. At $\beta=1$, the weights are assigned purely based on the distance set of current round. The simplified method strives to minimize the variance, aligning with the unbiased global model.

\subsection{Discussion and Limitation}\label{sec:discuss}
The key notations are summarized in \cref{tab:notation} and the pseudo-code of \ours{} is presented in \cref{alg:FedHEAL}.

\noindent \textbf{Comparison with Analogous Methods}. \qfedavg{} {\cite{qFedavg_ICLR19_fair}} and \fedce{} {\cite{FedCE_CVPR23_Fair}} increase weights for poor-performing clients based on single loss or accuracy metrics. However, increasing the weights of these clients does not guarantee significant performance improvement. \fael{} adjusts weights based on the domain diversity, which implies fitting gap between the global model and local models. It is a better way is to infer the effectiveness of weight modification based on the potential space for performance improvement. Similar work \fedfv{} {\cite{FedFV_IJCAI21}} alleviates model gradient conflicts. but it modifies gradients based on cosine similarity and gradient projection of the model, still allows less significant gradients to influence crucial ones. In contrast, \fphl{}, grounded in the observed \puc{} characteristics, selectively discards unimportant updates to safeguard the important ones. It demonstrates targeted conflict resolution, leading to better fairness and higher overall performance.

\noindent \textbf{Discussion on \fphl{}}. \fphl{} selectively discards unimportant parameters to reduce update conflicts, meaning it is particularly effective in large-scale FL systems where such conflicts are more pronounced. Similar with other methods aimed at Performance Fairness {\cite{qFedavg_ICLR19_fair,AFL_ICML19_agnostic}}, \fphl{} can increase the influence of clients with poorer performance, which may sometimes reduce the relative weight and performance of other clients. Yet, by discarding unimportant parameters uniformly, \fphl{} can diminish the adverse impact of both poorly performing client updates and better-performing clients to some extent. Consequently, while it significantly boosts the performance of the former, it may also help to prevent performance decline in the latter. So it achieves higher average accuracy across domains.

\noindent \textbf{Limitation}. Our method leverages the parameter update consistency and the fitting gap between the global model and local models to guide parameter aggregation and client aggregation weights. However, our method's performance is sensitive to the selected hyperparameters. When a hyperparameter is not selected properly, our method may become unstable. Additionally, our method is designed for scenarios where all clients share the same network architecture, so it may fail in cases where clients have different architectures and parameter update consistency cannot be assumed.

\setlength{\textfloatsep}{0.3cm}
\setlength{\floatsep}{0.3cm}
\begin{algorithm}[t]
\SetNoFillComment
\SetArgSty{textnormal}
\small{\KwIn{Communication rounds $T$, local epochs $\mathcal{K}$, number of participants $M$, $m^{th}$ participant private data $D_m$, private model $w_m$}}

\small{\KwOut{The final global model $w^{T}$}}
\BlankLine

\textbf{Server}: initialize the global model $w^0$ and {$L_m^0 = [0, 0, \ldots, 0]_{G}$}

\For {$t=0, 1, 2, ..., T-1$}{
    
    \textbf{Client}:
    
    \For {$m=1, 2, ..., M$ \textbf{in parallel}}{
    
    $w_m^t \leftarrow \mathcal{W}^t$
    
    \For {$k=1, 2, ..., \mathcal{K}$}{
        $w_m^t \leftarrow w_m^t - \eta \nabla \textbf{CE}(w_m^t,D_m)$
    }
    }

    $\Delta w_m^t \leftarrow w_{m}^{t} - \mathcal{W}^t$

    \textbf{Server}: 
    
    $q^t, L^t \leftarrow \textbf{FedHEAL}(L^{t-1})$
    
    
    \For {$i=1,2,\ldots,G$}{
         $\mathcal{W}_{i}^{t+1} = \mathcal{W}_{i}^t + \sum_{m=1}^M q_{m,i}^t \Delta w_{m,i}^t$
    }

}
$\textbf{FedHEAL}(L^{t-1})$ :

\For{$m=1,2,\ldots,M$}{
\For{$i=1,2,\ldots,G$}{
$l_{m,i} \leftarrow (l_{m,i}, \Delta w_{m,i}^t)$ in \cref{eq:update_incr}

$c_{m,i} \leftarrow (l_{m,i}, \Delta w_{m,i}^t)$ in \cref{eq:new_puc}
}
$d_m \leftarrow (c_{m,i}, \Delta w_{m,i}^t)$ in \cref{eq:new_dis}

}
\For{$m=1,2,\ldots,M$}{
$p_m^t, \Delta p_m^t \leftarrow (p_m^{t-1}, \Delta p_m^{t-1}, \Delta w_{m,i}^t, D^t, \beta) $ in \cref{eq:momentum}

\For{$i=1,2,\ldots,G$}{
$q_{m,i}^t \leftarrow (p_m^t, c_{m,i}^t)$ in \cref{eq:new_weight}
}
}
return $q^t$, $L^t$
\caption{FedHEAL}
\label{alg:FedHEAL}
\end{algorithm}

\section{Experiments}

\subsection{Experiment Details}
\label{sec:setup}
\noindent \textbf{Datasets}. We evaluate our methods on two multi-domain image classification tasks.
\begin{itemize}
	\setlength{\itemsep}{0pt}
	\setlength{\parsep}{-2pt}
	\setlength{\parskip}{-0pt}
	\setlength{\leftmargin}{-10pt}
        \item \digits{} {\cite{MNIST_IEEE98_digits,USPS_PAMI94_digits,SVHN_NeurIPS11_digits,SYN_arXiv18_digits}} includes four domains: MNIST, USPS, SVHN and SYN, each with 10 categories.
        \item \officecaltech{} {\cite{OfficeCaltech_CVPR12_OC}} includes four domains:  Caltech, Amazon, Webcam, and DSLR, each with 10 categories.
\end{itemize}
We allocate 20 clients for each task and distribute an equal number of clients to each domain. We randomly sample a certain proportion for each client from their datasets, based on task difficulty and task size. Specifically, we sample $1\%$ for \digits{} and $10\%$ for \officecaltech{}. We fix the seed to ensure reproduction of results. The example cases in each domain are presented in \cref{fig:examplecase}.

\begin{figure}[t]
\centering
\begin{minipage}[b]{0.45\linewidth}
    \includegraphics[width=\linewidth]{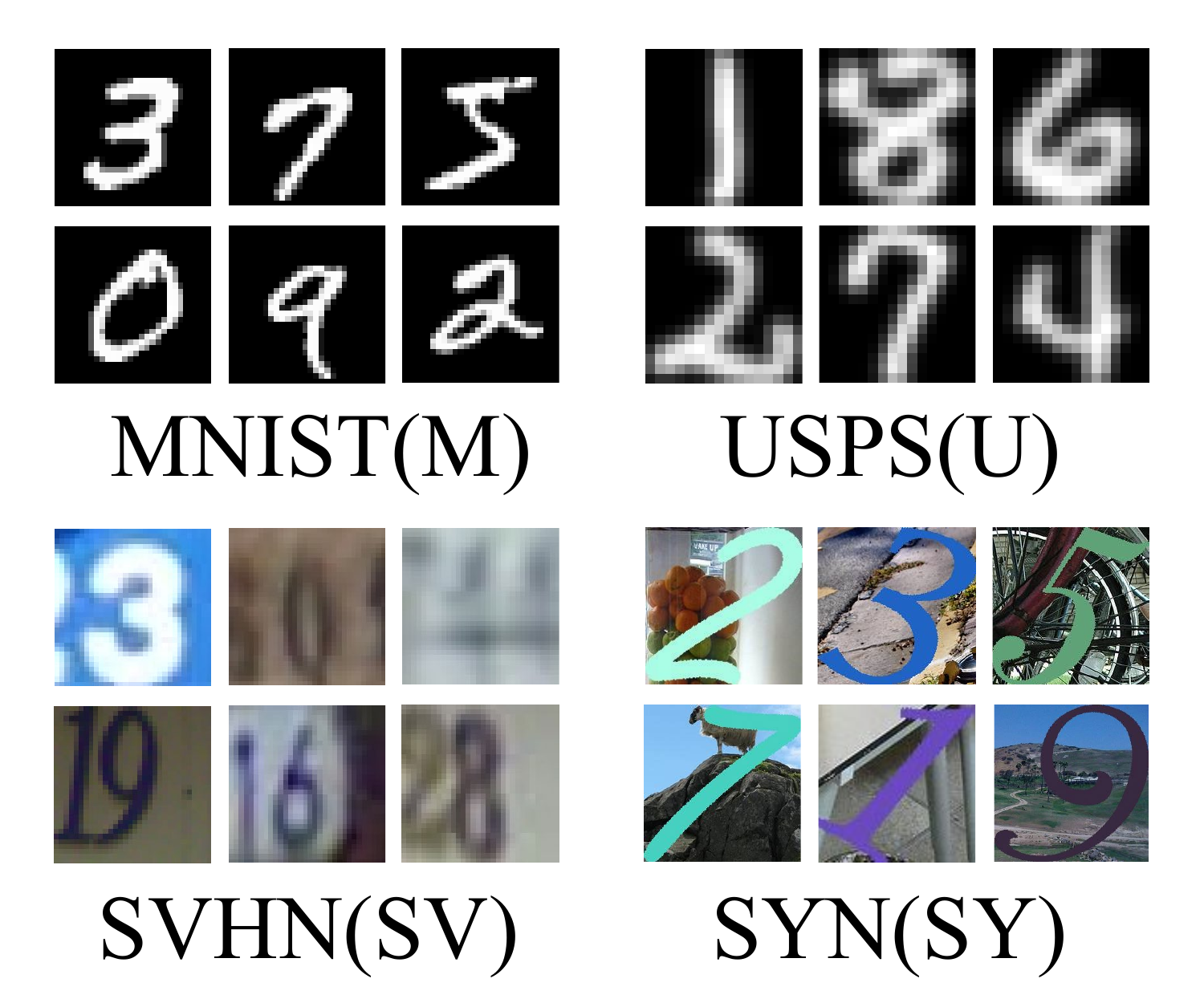}
    \vspace{-24pt} 
    \captionsetup{font=small}
    \caption*{\small{(a) \digits{}}}
    \label{fig:digits}
\end{minipage}
\hfill
\begin{minipage}[b]{0.45\linewidth}
    \includegraphics[width=\linewidth]{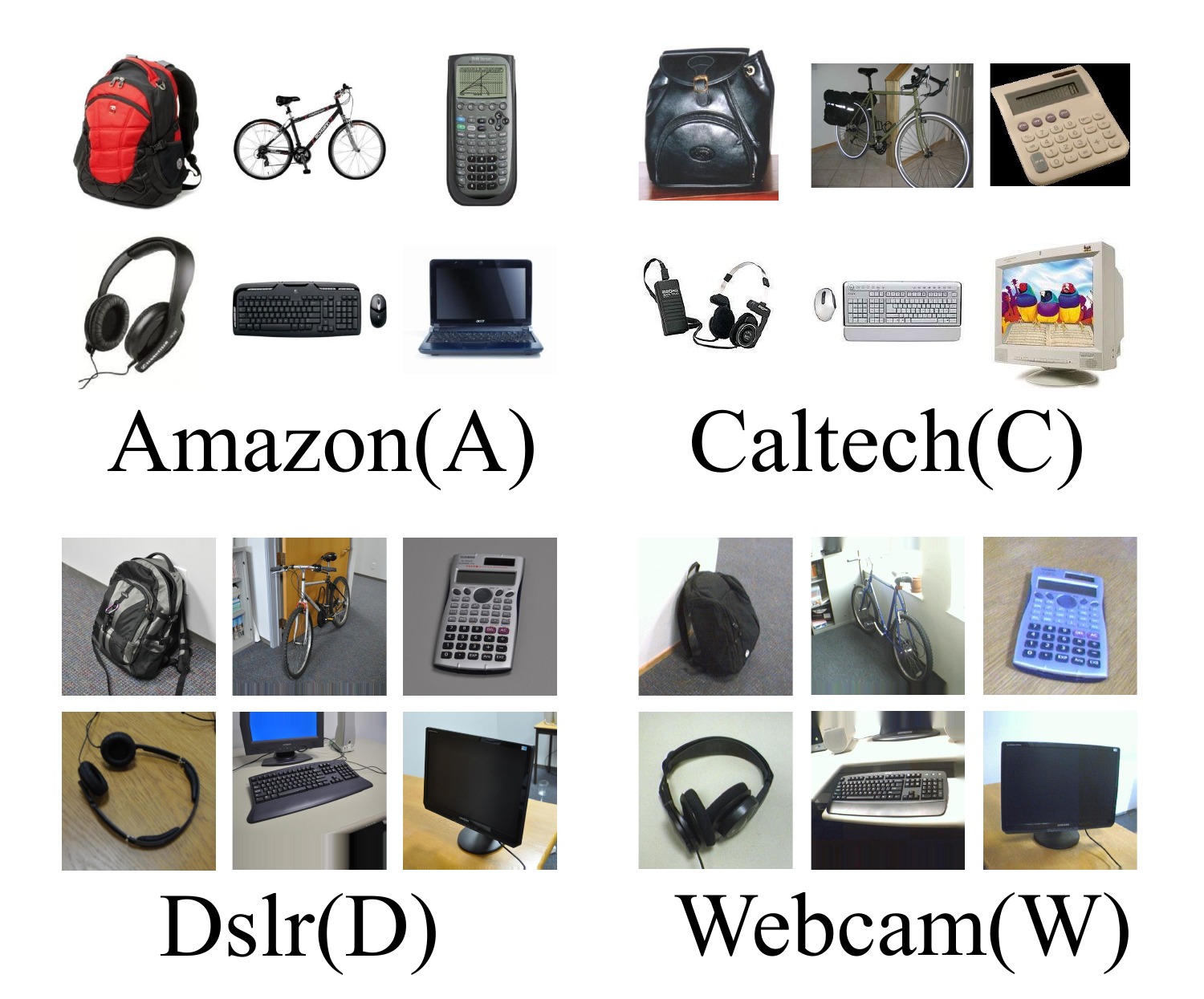}
    \vspace{-24pt} 
    \captionsetup{font=small}
    \caption*{\small{(b) \officecaltech{}}}
    \label{fig:officecaltech}
\end{minipage}
\vspace{-10pt} 
\caption{\small{
\textbf{Example cases} in \digits{} {\cite{MNIST_IEEE98_digits,USPS_PAMI94_digits,SVHN_NeurIPS11_digits,SYN_arXiv18_digits}}, \officecaltech{} {\cite{OfficeCaltech_CVPR12_OC}} tasks. Please see details in \cref{sec:setup}.
}}
\label{fig:examplecase}
\vspace{-7pt} 
\end{figure}

\noindent \textbf{Model}. For both classification tasks, we use \resnett{} {\cite{ResNet_CVPR16_res}} as the shared model architecture for training.

\begin{table*}[ht]\small
\setlength{\abovecaptionskip}{0cm}
\centering
    \resizebox{\linewidth}{!}{
        \renewcommand\arraystretch{1}
        \begin{tabular}{r||cccc|ccIcccc|cc}
		\hline\thickhline
            \rowcolor{lightgray}
		& \multicolumn{6}{cI}{\digits{}}&\multicolumn{6}{c}{\officecaltech{}}\\
        \cline{2-13}
        \rowcolor{lightgray}
    	\multirow{-2}{*}{Methods}& MNIST & USPS & SVHN  & SYN & AVG $\pmb{\uparrow}$ & STD $\pmb{\downarrow}$ & Amazon  & DSLR & Caltech  & Webcam & AVG $\pmb{\uparrow}$ & STD $\pmb{\downarrow}$  \\
    	\hline\hline
        \fedavg{} {\cite{FedAVG_AISTATS17_Communication}} 
        & 89.84 & 93.25 & 79.54 & 41.35 & 76.00 & 23.82 
        & 72.63 & 56.67 & 58.57 & 45.52 & 58.35 & 11.13 \\
        +\afl{} {\cite{AFL_ICML19_agnostic}} 
        & 90.59 & 95.83 & 75.13 & 44.42 & 76.49 & 23.12 
        & 64.21 & 65.37 & 57.50 & 48.28 & 58.83 & 7.84 \\
        +\qfedavg{} {\cite{qFedavg_ICLR19_fair}} 
        & 91.44 & 94.10 & 76.33 & 44.48 & 76.59 & 22.79 
        & 60.00 & 64.01 & 53.39 & 51.72 & 57.28 & 5.73 \\
        +\textbf{FedHEAL}  
        & 90.27 & 95.69 & 79.94 & 46.45 
        & \textbf{78.09}
        & \textbf{22.08} 
        & 67.90 & 66.00 & 59.28 & 66.21 
        & \textbf{64.85} 
        & \textbf{3.80}  \\
        \hline
        \fedprox{} {\cite{FedProx_MLSys20_LossOptimization}} 
        & 90.27 & 93.93 & 80.04 & 42.82 & 76.76 & 23.38 
        & 69.90 & 58.00 & 60.27 & 45.52 & 58.42 & 10.03 \\
        +\afl{} {\cite{AFL_ICML19_agnostic}} 
        & 92.86 & 96.17 & 74.47 & 42.22 & 76.43 & 24.72 
        & 68.10 & 62.67 & 59.29 & 52.41 & 60.62 & 6.57  \\
        +\qfedavg{} {\cite{qFedavg_ICLR19_fair}} 
        & 88.58 & 93.49 & 75.58 & 44.23 & 75.47 & 22.15 
        & 61.37 & 72.66 & 54.91 & 55.52 & 61.11 & 8.23 \\
        +\textbf{FedHEAL} 
        & 89.06 & 95.52 & 79.44 & 46.67 
        & \textbf{77.67} 
        & \textbf{21.70} 
        & 66.11 & 72.67 & 57.50 & 67.59 
        & \textbf{65.97} 
        & \textbf{6.30}   \\
        
        \hline
        \scaffold{} {\cite{SCAFFOLD_ICML20_stochastic}} 
        & 94.15 & 94.44 & 76.87 & 44.22 & 77.42 & 23.61 
        & 69.37 & 59.33 & 59.55 & 46.21 & 58.62 & 9.50 \\
        +\afl{} {\cite{AFL_ICML19_agnostic}} 
        & 91.77 & 96.05 & 78.60 & 46.39 & 78.20 & 22.47 
        & 66.42 & 63.33 & 59.11 & 49.31 & 59.54 & 7.45  \\
        +\qfedavg{} {\cite{qFedavg_ICLR19_fair}} 
        & 87.73 & 94.59 & 74.00 & 43.76 & 75.02 & 22.53 
        & 61.79 & 73.33 & 55.18 & 55.86 & 61.54 & 8.40 \\
        +\textbf{FedHEAL}  
        & 92.68 & 96.25 & 78.54 & 47.72 
        & \textbf{78.80}  
        & \textbf{22.08}  
        & 64.11 & 67.99 & 55.18 & 62.41 
        & \textbf{62.42}  
        & \textbf{5.37}  \\
        
        \hline
        \moon{} {\cite{MOON_CVPR21_ContrastiveL}} 
        & 90.46 & 92.65 & 80.48 & 40.58 & 76.04 & 24.23 
        & 74.00 & 59.33 & 60.63 & 46.90 & 60.21 & 11.08 \\
        +\afl{} {\cite{AFL_ICML19_agnostic}} 
        & 91.25 & 96.03 & 75.31 & 44.34 & 76.73 & 23.34 
        & 66.74 & 67.33 & 60.80 & 55.17 & 62.51 & 5.71  \\
        +\qfedavg{} {\cite{qFedavg_ICLR19_fair}} 
        & 90.43 & 94.84 & 76.48 & 43.95 & 76.42 & 23.02 
        & 64.32 & 65.33 & 54.28 & 61.03 & 61.24 & 4.99 \\
        +\textbf{FedHEAL}  
        & 91.34 & 94.94 & 81.32 & 44.96 
        & \textbf{78.14} 
        & \textbf{22.86}  
        & 67.68 & 65.33 & 59.11 & 64.14 
        & \textbf{64.07}  
        & \textbf{3.62}   \\
        
        \hline
        \feddyn{} {\cite{FedDyn_ICLR21_Regularization}} 
        & 91.23 & 92.36 & 80.15 & 41.55 & 76.32 & 23.83 
        & 71.16 & 62.00 & 59.20 & 48.62 & 60.24 & 9.28 \\
        +\afl{} {\cite{AFL_ICML19_agnostic}} 
        & 92.11 & 96.10 & 71.46 & 41.52 & 75.30 & 24.97
        & 70.10 & 58.67 & 59.82 & 51.03 & 59.91 & 7.84  \\
        +\qfedavg{} {\cite{qFedavg_ICLR19_fair}} 
        & 92.53 & 95.17 & 76.37 & 44.75 & 77.20 & 23.18 
        & 62.10 & 67.33 & 54.82 & 56.21 & 60.12 & 5.76 \\
        +\textbf{FedHEAL}  
        & 89.87 & 95.00 & 80.18 & 44.23 
        & \textbf{77.32} 
        & \textbf{22.90}  
        & 67.47 & 60.66 & 59.02 & 54.83 
        & \textbf{60.50}  
        & \textbf{5.26}   \\
        \hline
        
        \fedproc{} {\cite{FedProc_arXiv21}} 
        & 91.86 & 91.16 & 78.54 & 39.87 & 75.36 & 24.44 
        & 60.21 & 46.00 & 55.98 & 46.90 & 52.27 & 6.95 \\
        +\afl{} {\cite{AFL_ICML19_agnostic}} 
        & 87.85 & 94.28 & 78.52 & 41.54 & 75.55 & 23.58 
        & 52.63 & 52.67 & 55.09 & 43.45 & 50.96 & 5.14  \\
        +\qfedavg{} {\cite{qFedavg_ICLR19_fair}} 
        & 92.09 & 92.09 & 74.97 & 45.21 & 76.15 & 22.17 
        & 65.79 & 42.01 & 55.80 & 50.69 & 53.57 & 9.94 \\
        +\textbf{FedHEAL} 
        & 94.23 & 92.93 & 81.43 & 48.67 
        & \textbf{79.31}  
        & \textbf{21.22}   
        & 67.58 & 66.00 & 56.79 & 61.38 
        & \textbf{62.94}  
        & \textbf{4.87}  \\
        \hline
        \fedproto{} {\cite{FedProto_AAAI22_PrototypeLoss}} 
        & 89.99 & 92.90 & 81.09 & 40.93 & 76.23 & 24.06 
        & 71.48 & 42.67 & 62.23 & 60.34 & 59.18 & 12.04 \\
        +\afl{} {\cite{AFL_ICML19_agnostic}} 
        & 85.27 & 92.90 & 67.16 & 42.36 & 71.92 & 22.47 
        & 70.74 & 56.67 & 57.77 & 79.65 & 66.21 & 11.01  \\
        +\qfedavg{} {\cite{qFedavg_ICLR19_fair}} 
        & 93.35 & 94.92 & 77.08 & 46.31 & 77.91 & 22.56 
        & 72.74 & 54.67 & 64.20 & 82.76 & 68.59 & 11.99 \\
        +\textbf{FedHEAL} 
        & 88.49 & 94.62 & 81.39 & 48.46 
        & \textbf{78.24} 
        & \textbf{20.58} 
        & 75.68 & 76.00 & 65.18 & 80.34 
        & \textbf{74.30} 
        & \textbf{6.44}  \\
        \hline\thickhline
        
        \end{tabular}}
	\vspace{3pt}
    \caption{Comparison of Average Accuracy(\textbf{AVG}) and Standard Deviation(\textbf{STD}) with \afl{} {\cite{AFL_ICML19_agnostic}} and \qfedavg{} {\cite{qFedavg_ICLR19_fair}}. See details in \cref{sec:sota}.}
	\label{tab:acc}
 \vspace{-18pt}
\end{table*}

\noindent \textbf{Comparison Methods}. We compare \ours{} with FL baseline \fedavg{} {\cite{FedAVG_AISTATS17_Communication}} and existing solutions for Performance Fairness: \afl{} {\cite{AFL_ICML19_agnostic}}, \qfedavg{} {\cite{qFedavg_ICLR19_fair}} (both integrable), \fedfv{} {\cite{FedFV_IJCAI21}}, \ditto{} {\cite{Ditto_ICML21_Ditto}} (two independent methods, with personalized models aggregated into global model for \ditto{}).

\noindent \textbf{Implementation Details}. All methods are implemented with the same settings. We set the communication rounds to 200 and the local epoch to 10. We use SGD as the optimizer with a learning rate of 0.001. Its weight decay is $1e-5$ and momentum is $0.9$. The training batch size is $64$ for \digits{} and $16$ for \officecaltech{}. The hyper-parameter setting for \ours{} presents in the \cref{sec:hyperparam}.

\noindent \textbf{Evaluation Metrics}.
Following {\cite{qFedavg_ICLR19_fair}}, we utilize the Top-1 accuracy and the standard deviation of accuracy across multi-domains as evaluation metrics. A smaller standard deviation indicates better Performance Fairness across different domains. We use the average results from the last five rounds accuracy and variance as the final performance.

\begin{table}[ht]
\vspace{-6pt}
\centering
\small 
\setlength{\tabcolsep}{2pt} 
\scalebox{0.96}{
\begin{tabular}{c||cccc|cc}
\hline \thickhline
\rowcolor{lightgray}
& \multicolumn{6}{c}{\digits{}}\\
\cline{2-7} 
\rowcolor{lightgray}
\multirow{-2}{*}{Methods}  
& \mnist{} & \usps{} & \svhn{} & \syn{} & AVG $\pmb{\uparrow}$ & STD $\pmb{\downarrow}$\\
\hline\hline
\ditto{} {\cite{Ditto_ICML21_Ditto}} 
& 90.59 & 92.98 & 79.20 & 41.89 & 76.16 & 23.62 \\
\fedfv{} {\cite{FedFV_IJCAI21}} 
& 91.76 & 94.70 & 77.26 & 44.14 & 76.97 & 23.17 \\
\textbf{\ours{}} 
& 90.27 & 95.69 & 79.94 & 46.45 & \textbf{78.09} & \textbf{22.08} \\
\hline \thickhline
\rowcolor{lightgray}
& \multicolumn{6}{c}{\officecaltech{}}\\
\cline{2-7} 
\rowcolor{lightgray}
\multirow{-2}{*}{Methods}  
& \amazon{} & \dslr{} & \caltech{} & \webcam{} & AVG $\pmb{\uparrow}$ & STD $\pmb{\downarrow}$\\
\hline\hline

\ditto{} {\cite{Ditto_ICML21_Ditto}} 
& 58.00 & 70.00 & 56.25 & 63.45 & 61.92 & 6.20 \\
\fedfv{} {\cite{FedFV_IJCAI21}} 
& 62.95 & 71.33 & 55.36 & 60.00 & 62.41 & 6.72 \\
\textbf{\ours{}} 
&  67.90 & 66.00 & 59.28 & 66.21 & \textbf{64.85} & \textbf{3.80} \\
\hline\thickhline
\end{tabular}

}
\vspace{-5pt}
\caption{Comparison of Average Accuracy(\textbf{AVG}) and Standard Deviation(\textbf{STD}) with \ditto{} {\cite{Ditto_ICML21_Ditto}} and \fedfv{} {\cite{FedFV_IJCAI21}}. Please refer to \cref{sec:sota} for detailed discussion.}
\label{tab:sota2}
\vspace{-8pt}
\end{table}

\subsection{Diagnostic Analysis}
\label{sec:hyperparam}
\noindent \textbf{Hyper-parameter Study}. We show the impact of the hyper-parameters $ \tau $ (\cref{eq:new_weight}) and $ \beta $ (\cref{eq:momentum}) on the performance in \cref{tab:beta} and \cref{fig:tau}, in \digits{}, optimal performance is achieved when $\beta=0.4$ and $\tau=0.3$. Similar experiments on \officecaltech{} yields $ \beta=0.4 $ and $ \tau=0.4 $ as the best settings. We use these hyper-parameters by default in subsequent experiments. 

\noindent \textbf{Ablation Study}. To provide a comprehensive analysis of the effectiveness of \fphl{} and \fael{}, we carried out an ablation study on both the \digits{} and \officecaltech{} in \cref{tab:ablation}. They contribute positively to the performance enhancement and their combination results in optimal performance. 

\begin{figure}[t]
\centering
\includegraphics[trim=50 500 60 50, clip, width=\linewidth]{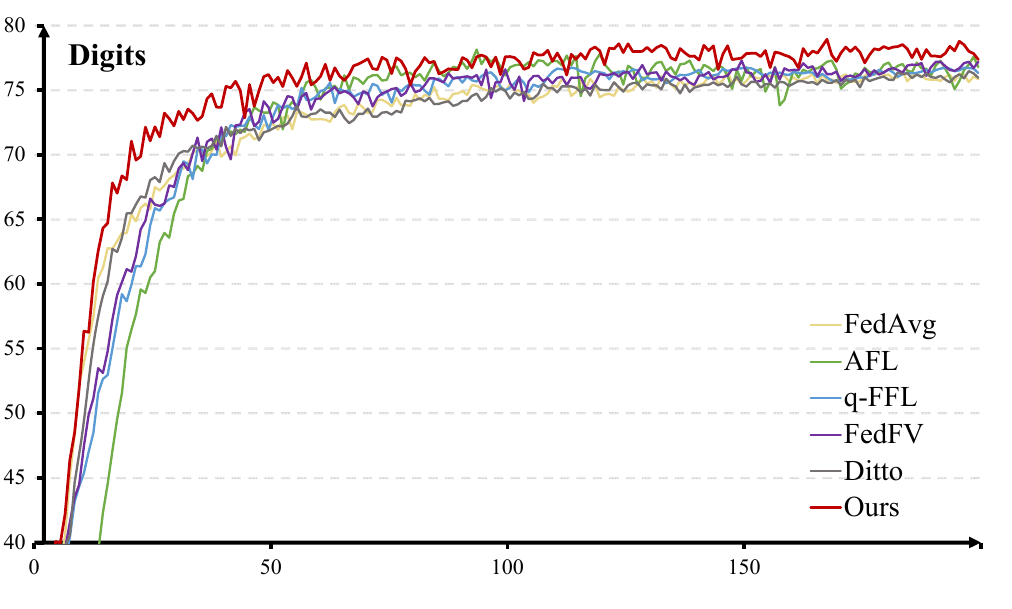}
\vspace{-20pt}
\captionsetup{font=small}
\caption{\small{
\textbf{Comparison of convergence of average accuracy} with counterparts on \digits{}. Please see details in \cref{sec:sota}.
}}
\label{fig:compare_sota}
\vspace{-2pt}
\end{figure}

\begin{table}[t]
\centering
\small 
\begin{tabular}{c||cccccc}
\hline \thickhline
\rowcolor{mygray}
$\beta$ & 0.0 & 0.2 & 0.4 & 0.6 & 0.8 & 1.0 \\
\hline\hline
AVG $\pmb{\uparrow}$ & 76.00 & 76.88 & \textbf{77.20} & 76.83 & 76.96 & 77.13 \\
STD $\pmb{\downarrow}$ & 23.82 & 22.70 & \textbf{22.64} & 22.83 & 22.71 & 22.70 \\
\hline\thickhline
\end{tabular}
 \vspace{-3mm}
\caption{\textbf{Hyper-parameter study} with different $\beta$ (\cref{eq:momentum}) on \digits{} datasets. See details in \cref{sec:hyperparam}.}
\label{tab:beta}
\end{table}

\begin{figure}[t]
\centering
\begin{minipage}[b]{0.49\linewidth}
    \includegraphics[trim=0 40 60 4, clip, width=\linewidth]{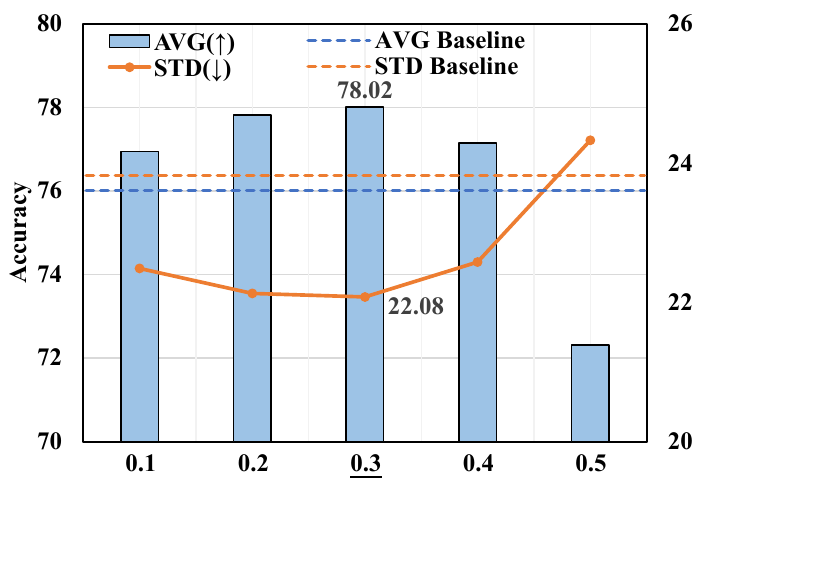}
    \vspace{-2em} 
    \captionsetup{font=small}
    \caption*{\small{(a) \digits{}}}
    \label{fig:digits_tau}
\end{minipage}
\hfill
\begin{minipage}[b]{0.49\linewidth}
    \includegraphics[trim=5 40 55 4, clip, width=\linewidth]{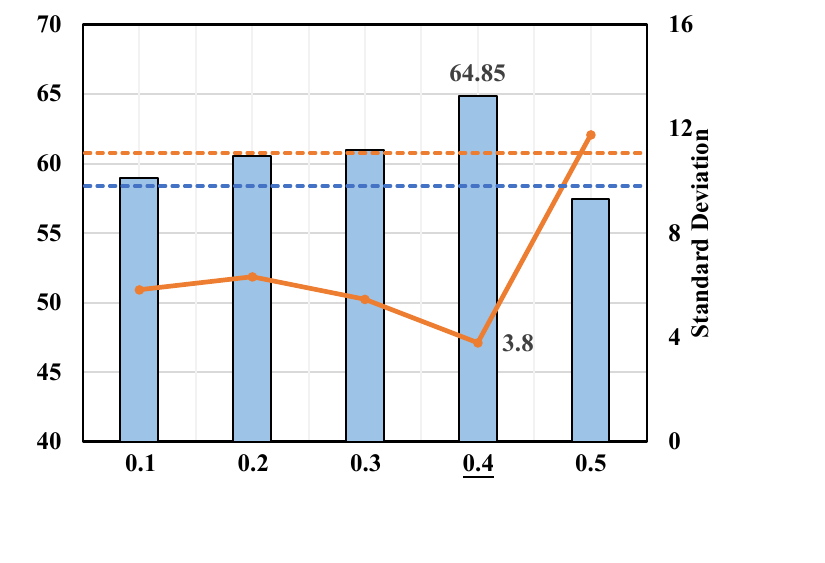}
    \vspace{-2em} 
    \captionsetup{font=small}
    \caption*{\small{(b) \officecaltech{}}}
    \label{fig:officecaltech_tau}
\end{minipage}
\vspace{-8pt} 
\caption{\small{
\textbf{Hyper-parameter study} with variant $\tau$ (\cref{eq:incre_prop}) when fix $\beta=0.4$. Please see details in \cref{sec:hyperparam}.
}}
\label{fig:tau}
\vspace{-5pt} 
\end{figure}

\noindent \textbf{Compatibility Study}. To validate the compatibility of \ours{}, we compared the results of several widely-adopted FL methods, \fedavg{} {\cite{FedAVG_AISTATS17_Communication}}, \fedprox{} {\cite{FedProx_MLSys20_LossOptimization}}, \scaffold{} {\cite{SCAFFOLD_ICML20_stochastic}}, \moon{} {\cite{MOON_CVPR21_ContrastiveL}}, \feddyn{} {\cite{FedDyn_ICLR21_Regularization}}, \fedproc{} {\cite{FedProc_arXiv21}}, \fedproto{} {\cite{FedProto_AAAI22_PrototypeLoss}} without and with \ours{}. The results are shown in \cref{tab:acc}. They reveal tangible benefits offered by our system, \ie, \fedproto{} {\cite{FedProto_AAAI22_PrototypeLoss}} with \ours{} achieves $5.60\%$ reduction in STD and $15.12\%$ increase in AVG on \officecaltech{}. We plot the differences in convergence between the benchmark without and with \ours{} in \cref{fig:integration}, demonstrating faster convergence and higher accuracy of \ours{}.

\subsection{Comparison to State-of-the-Arts}\label{sec:sota}
The \cref{tab:acc} and \cref{tab:sota2} shows the accuracy and standard deviation at the end of communication with SOTAs that address Performance Fairness in FL. The results depict that our method outperforms counterparts in both standard deviation and mean accuracy. This demonstrates that \ours{} achieves better Performance Fairness and further improves accuracy across multiple domains. We plot the average accuracy at each epoch in \cref{fig:compare_sota}, which illustrates the faster convergence of \ours{}.

\begin{figure}[t]
\centering
\includegraphics[trim=50 500 60 50, clip, width=\linewidth]{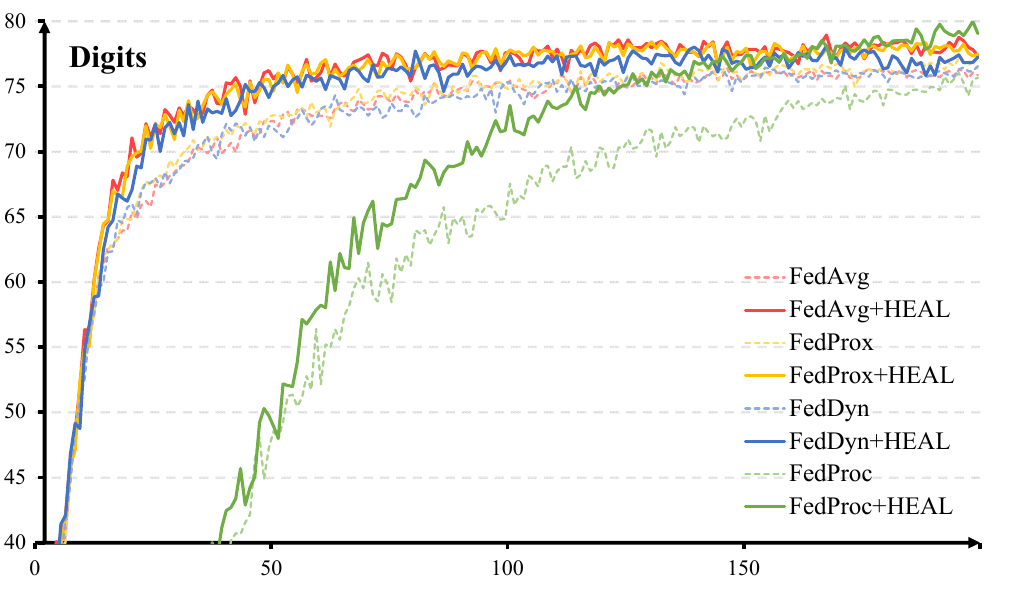}
\vspace{-20pt}
\captionsetup{font=small}
\caption{\small{
\textbf{Comparison of convergence of average accuracy} with and without the integration of \ours{}, across selected FL methods. Please see details in \cref{sec:hyperparam}.}}
\vspace{-5pt}
\label{fig:integration}
\end{figure}

\begin{table}[t]
\centering
\small 
\setlength{\tabcolsep}{2pt} 
\scalebox{0.94}{
\begin{tabular}{cc||cccc|cc}
\hline \thickhline
\rowcolor{lightgray}
&& \multicolumn{6}{c}{\digits{}}\\
\cline{3-8} 
\rowcolor{lightgray}
\multirow{-2}{*}{\fphl{}} & \multirow{-2}{*}{\fael{}} 
& \mnist{} & \usps{} & \svhn{} & \syn{} & AVG $\pmb{\uparrow}$ & STD $\pmb{\downarrow}$\\
\hline\hline
& 
& 89.84 & 93.25 & 79.54 & 41.35 & 76.00 & 23.82\\
\ding{51} &  
& 92.19 & 95.32 & 76.32 & 44.37 & 77.05 & 23.32\\
& \ding{51} 
& 90.05 & 95.16 & 78.76 & 44.83 & 77.20 & 22.64 \\
\ding{51} & \ding{51}
& 90.27 & 95.69 & 79.94 & 46.45 & \textbf{78.09} & \textbf{22.08} \\
\hline \thickhline
\rowcolor{lightgray}
&& \multicolumn{6}{c}{\officecaltech{}}\\
\cline{3-8} 
\rowcolor{lightgray}
\multirow{-2}{*}{\fphl{}} & \multirow{-2}{*}{\fael{}} 
& \amazon{} & \dslr{} & \caltech{} & \webcam{} & AVG $\pmb{\uparrow}$ & STD $\pmb{\downarrow}$\\
\hline\hline
& 
& 72.63 & 56.67 & 58.57 & 45.52 & 58.35 & 11.13\\
\ding{51} &  
& 68.42 & 66.00 & 57.95 & 66.55 & 64.73 & 4.64\\
& \ding{51} 
& 66.73 & 63.33 & 57.59 & 53.10 & 60.19 & 6.05\\
\ding{51} & \ding{51}
&  67.90 & 66.00 & 59.28 & 66.21 & \textbf{64.85} & \textbf{3.80} \\
\hline\thickhline
\end{tabular}
}
 \vspace{-3mm}
\caption{\textbf{Ablation study} on \digits{} and \officecaltech{}. Pleaese refer to \cref{sec:hyperparam} for detailed discussion.}
\label{tab:ablation}
\end{table}

\section{Conclusion}
In this paper, we address Performance Fairness in federated learning with domain skew by tackling parameter update conflicts and model aggregation bias. We discover a property in federated learning which we term \textit{Parameter Update Consistency}. Leveraging this characteristic, we propose a simple yet effective approach. By discarding unimportant parameters, \ours{} alleviates parameter update conflicts for poor-performing clients. Moreover, considering domain diversity, we reduce the variance of distances between the global model and local models, addressing model aggregation bias. Extensive experiments demonstrate the effectiveness and compatibility of \ours{}. We believe that this newly discovered property and our work will offer fresh research directions and insights for the community.

\noindent \textbf{Acknowledgement.}
This work is supported by National Natural Science Foundation of China under Grant (62361166629, 62176188, 62272354).

{
    \small
    \bibliographystyle{ieeenat_fullname}
    \bibliography{main}
}


\end{document}